\pdfoutput=1

\documentclass[11pt]{article}

\usepackage{acl}

\usepackage{times}
\usepackage{latexsym}
\usepackage{amsmath}
\usepackage{amssymb}
\usepackage{booktabs}
\usepackage{multirow}
\usepackage{multicol}
\usepackage{makecell}
\usepackage{CJKutf8}
\usepackage{graphicx}
\usepackage{subfigure}
\usepackage{url}
\usepackage[T1]{fontenc}

\usepackage[utf8]{inputenc}

\usepackage{microtype}

\usepackage{inconsolata}

\usepackage{graphicx}

\title{Learning to Solve Domain-Specific Calculation Problems with Knowledge-Intensive Programs Generator}

\author{\textbf{Chengyuan Liu\textsuperscript{1,2}\footnotemark[2],\quad Shihang Wang\textsuperscript{2},\quad Lizhi Qing\textsuperscript{2}}\\ \textbf{Jun Lin\textsuperscript{2},\quad Ji Zhang\textsuperscript{2},\quad Fei Wu\textsuperscript{1},\quad Kun Kuang\textsuperscript{1}\footnotemark[1]}\\
\textit{\{liucy1,wufei,kunkuang\}@zju.edu.cn}, \\ \textit{\{wangshihang.wsh,yekai.qlz,linjun.lj,zj122146\}@alibaba-inc.com}
\\
\textit{\small \textsuperscript{1}College of Computer Science and Technology, Zhejiang University,}
\textit{\small \textsuperscript{2}Tongyi Lab, Alibaba Group}\\
}

\begin{document}
\maketitle
\renewcommand{\thefootnote}{\fnsymbol{footnote}}
\footnotetext[1]{Corresponding author.}
\footnotetext[2]{This work was done when Chengyuan Liu interned at Alibaba.}
\begin{abstract}
Domain Large Language Models (LLMs) are developed for domain-specific tasks based on general LLMs.
But it still requires professional knowledge to facilitate the expertise for some domain-specific tasks.
In this paper, we investigate into knowledge-intensive calculation problems. We find that the math problems to be challenging for LLMs, when involving complex domain-specific rules and knowledge documents, rather than simple formulations of terminologies. Therefore, we propose a pipeline to solve the domain-specific calculation problems with \textbf{K}nowledge-\textbf{I}ntensive \textbf{P}rograms \textbf{G}enerator more effectively, named as \textbf{KIPG}. It generates knowledge-intensive programs according to the domain-specific documents. For each query, key variables are extracted, then outcomes which are dependent on domain knowledge are calculated with the programs. By iterative preference alignment, the code generator learns to improve the logic consistency with the domain knowledge. Taking legal domain as an example, we have conducted experiments to prove the effectiveness of our pipeline, and extensive analysis on the modules. We also find that the code generator is also adaptable to other domains, without training on the new knowledge.

\end{abstract}

\section{Introduction}

Large Language Models (LLMs) \cite{brown2020languagemodelsfewshotlearners, anil2023palm2technicalreport, yang2023baichuan2openlargescale, touvron2023llama2openfoundation} have exhibited outstanding results on several tasks, covering both language understanding and generation, even without additional training. Based on which, domain-specific LLMs \cite{zhou2024lawgptchineselegalknowledgeenhanced, yang2023fingptopensourcefinanciallarge, zhang2023huatuogpttaminglanguagemodel} are developed via techniques such as continual pre-training and supervised fine-tuning (SFT).
Although the LLMs are enhanced by domain corpus, it still requires professional knowledge to facilitate the expertise for some domain-specific tasks \cite{wisdomInterrogatory, ma2024sciagenttoolaugmentedlanguagemodels, nay2023largelanguagemodelstax}.
In this paper, we mainly study at the domain-specific QA tasks involving math calculation.

\begin{figure}
    \centering
    \includegraphics[width=\linewidth]{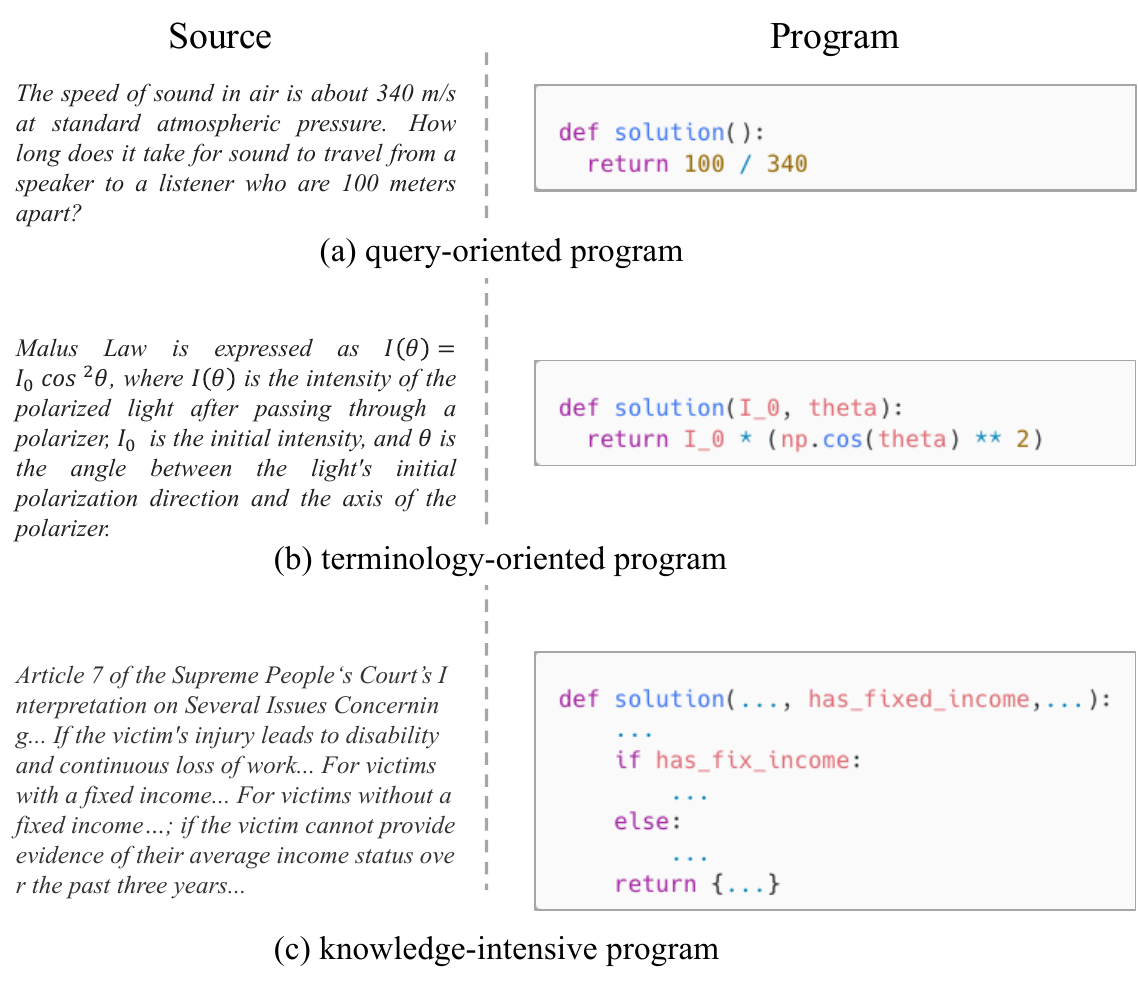}
    \caption{Comparison between program generation paradigms from different sources. (a) Query-oriented programs are generated to solve a specific query. (b) Terminology-oriented programs describe domain-specific concepts with several formulations, source from definitions of terminologies. (c) Knowledge-intensive programs follow more complex instructions with domain-specific conditions and rules.}
    \label{fig:comp}
\end{figure}

Since LLMs are not good at numeric calculation in auto-regression, researchers adopt algebraic expressions \cite{imani-etal-2023-mathprompter} and programs \cite{zhao2024financemathknowledgeintensivemathreasoning} to calculate precise numeric answers. The approaches can be classified by the source of generation, as shown in Figure \ref{fig:comp}. \citet{zhao2024financemathknowledgeintensivemathreasoning} conduct analysis on Program of Thought (PoT) for financial domain. The programs are generated to answer the query directly. We call them the \textbf{query-oriented programs}, as in Figure \ref{fig:comp} (a). To introduce external knowledge, \citet{ma2024sciagenttoolaugmentedlanguagemodels} extracts several definitions of terminologies from scientific domain as executable tools in Python language, and builds a dataset to train the LLM's ability to call the tools. As shown in Figure \ref{fig:comp} (b), the programs describe domain-specific concepts and terminologies and return the value of the concept, named \textbf{terminology-oriented programs} in this paper.

\begin{figure}
    \centering
    \includegraphics[width=\linewidth]{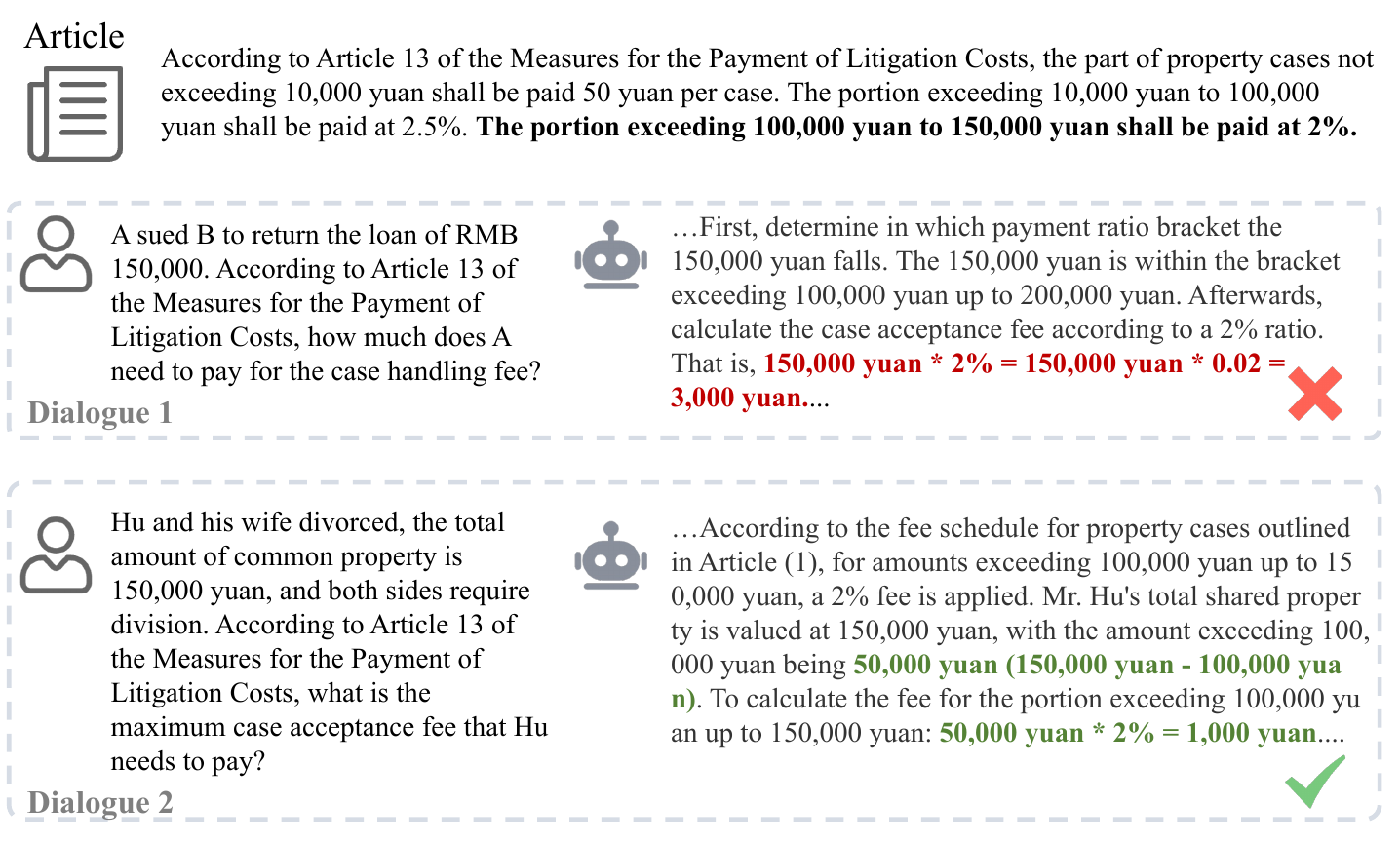}
    \caption{For the same article, the response may follow different calculation logic given different queries. Note that we only present part of the responses containing the logic concerned, i.e., ``\textit{The portion exceeding 100,000 yuan to 150,000 yuan shall be paid at 2\%.}''.}
    \label{fig:badcase}
\end{figure}

From a pilot study, we find that it can be hard for LLMs to strictly follow the domain knowledge represented in documents involving instructions with complex conditions and constraints, rather than simple formulations. As illustrated in Figure \ref{fig:badcase}, the understanding to the article rules shifts with different queries, thus the LLM may generate unstable answers. Therefore, we introduce \textbf{knowledge-intensive programs}, as in Figure \ref{fig:comp} (c), which may contain hierarchical conditions involving diverse variables. Additionally, knowledge-intensive programs return all potential results from the knowledge, instead of a single scalar as query/terminology-oriented programs. In this way, the bias caused by queries can be removed by representing the essential logic of the article as knowledge-intensive program.

In this paper we design a pipeline to solve the domain-specific calculation problems with \textbf{K}nowledge-\textbf{I}ntensive \textbf{P}rograms \textbf{G}enerator step by step, called \textbf{KIPG}. For each query, key variables are extracted, then outputs dependent on domain knowledge are calculated with the programs. Given the outputs, the LLM is prompted to conclude the answer. We also propose a method to train a code generator, thus the knowledge-intensive programs can be generated automatically when facing new knowledge documents.

Additionally, we have constructed a QA dataset involving numeric calculation in legal domain with the help of human labor from legal experts. Various law articles are included covering a diversity of cases. From the experiments, our proposed method significantly outperforms the baselines for most of the cases. Given the gold article, KIPG surpasses baselines by a margin. We also find that the learned code generator can be generalized to other domains. In summary, our contributions are three folds:
\begin{itemize}
    \item We have constructed a numeric calculation dataset involving domain-specific knowledge with complex conditions and instructions, rather than only single definitions of domain-specific terminologies.
    \item We introduce a pipeline, KIPG, to solve the domain-specific math word problems with knowledge-intensive programs generator, which learns to write the programs automatically according to the domain knowledge.
    \item We conduct experiments and analysis from multiple perspectives to demonstrate the effectiveness of our framework.
\end{itemize}

\begin{figure*}
    \centering
    \includegraphics[width=\linewidth]{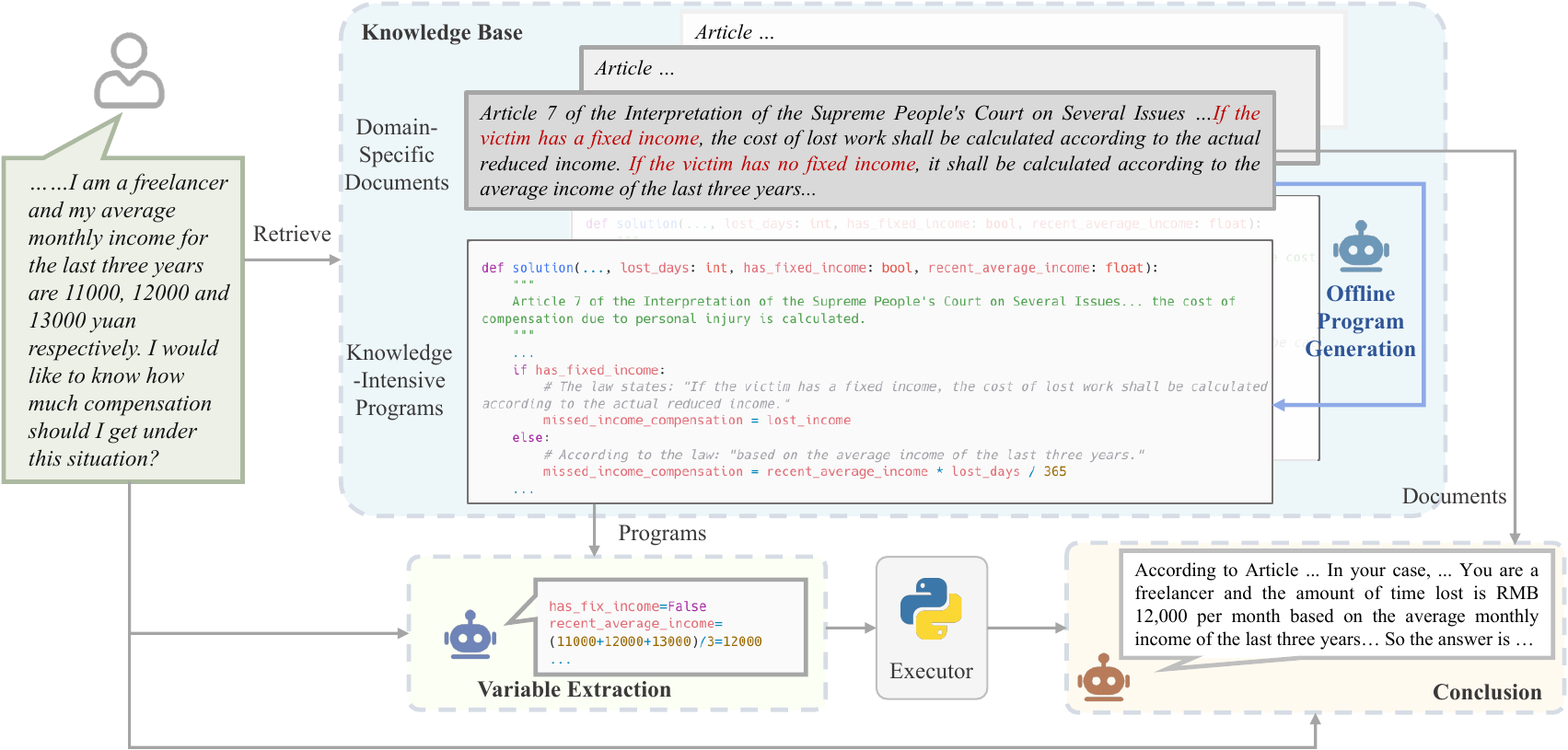}
    \caption{Framework of KIPG. The generator LLM writes programs for each domain-specific document. Given a query, KIPG extracts key variables and executes the programs to calculate the outcomes dependent on domain-specific knowledge step by step, finally concludes the answer.}
    \label{fig:framework}
\end{figure*}

\section{Related Work}

\paragraph{Domain-specific LLMs}
Domain-specific LLMs are developed based on general LLMs for domain-specific tasks. With the help of continual per-training \cite{ke2023continualpretraininglanguagemodels, yıldız2024investigatingcontinualpretraininglarge}, knowledge from various domains is injected into the parameters of LLMs. \citet{zhou2024lawgptchineselegalknowledgeenhanced} introduced LawGPT, the first open-source model specifically designed for Chinese legal applications. \citet{yang2023fingptopensourcefinanciallarge} presented an open-source large language model, FinGPT, for the finance sector. \citet{zhang2023huatuogpttaminglanguagemodel} presented HuatuoGPT, another LLM for medical consultation.

\paragraph{Math Problems for LLMs}
Auto-regression cannot ensure the correctness of numeric calculation. Researchers have paid much efforts to improve the calculation and reasoning abilities of LLMs. GSM8K \cite{cobbe2021trainingverifierssolvemath} is one of the most popular datasets for math calculation. \citet{wei2023chainofthoughtpromptingelicitsreasoning} introduced Chain-of-Thought (CoT) to elicit reasoning and calculation abilities of LLMs by straightforward prompts. \citet{imani-etal-2023-mathprompter} proposed MathPrompter, a technique that improves performance of LLMs on arithmetic problems along with increased reliance in the predictions. \citet{kim2023languagemodelssolvecomputer} showed that a LLM agent can execute computer tasks guided by natural language using a simple prompting scheme where the agent Recursively Criticizes and Improves its output (RCI).

\paragraph{Domain-specific Math Word Problems}
\citet{zhao2024financemathknowledgeintensivemathreasoning} explored using query-oriented programs for solving knowledge-intensive math reasoning problems.
\citet{eedp} introduced a novel prompting technique for financial tabular question-answering tasks involving math calculation.
\citet{nay2023largelanguagemodelstax} explored LLM capabilities in applying tax law to math calculation tasks.
\citet{ma2024sciagenttoolaugmentedlanguagemodels} used terminology-oriented programs as tools for scientific problem solving, which mainly describe simple formulation of terminologies and concepts. While we utilize knowledge-intensive programs in this paper, which involve more complex conditions and instructions.

\section{Dataset Construction}

For experimental analysis, we construct a dataset in legal domain to evaluate the calculation ability of LLMs given domain-specific knowledge describing complex instructions. We also collect various knowledge to build another dataset in medical domain as a testset, to assess the generalization. Details of the dataset construction are described in Appendix \ref{app:guide dataset}.

\begin{table}[th]
    \centering
    \scriptsize
    \begin{tabular}{clccc}
    \toprule
    Domain & Type & \# SubType & \# Article & \# Instance\\
    \midrule
    \multirow{5}{*}{Legal} & Compensation & 12 & 17 & 220 \\
    & Tax & 11  & 6 & 216\\
    & Other Fees & 8 & 10 & 201\\
    & Penalties & 2 & 552 & 1277\\
    & Traffic Violations & 1 & 6 & 136 \\
    \midrule
    \multirow{3}{*}{Medical} & Term & 1 & 4 & 153\\
    & indicator & 1 & 6 & 68\\
    & Medicine & 1 & 5 & 139\\
    \bottomrule
    \end{tabular}
    \caption{Statistics of the constructed dataset.}
    \label{tab:data info}
\end{table}

\paragraph{Knowledge Selection}
To improve the challenge of the task, we focus on practical domain knowledge, which mostly contains detailed instructions for different cases. 
It is non-trivial for LLMs to strictly follow the complex knowledge document to perform calculation given queries, especially considering that it requires domain knowledge to identify the corresponding conditions.
For legal domain, we have engaged professional lawyers with expertise in Chinese law to identify satisfactory law articles that involve mathematical computations. For medical domain, we search for medication instructions and indicators involving computations from the Internet to ensure the correctness of the documents. Some statistics are provided in Table \ref{tab:data info}. 

\paragraph{Query and Responses}
The queries should ask a specific question and expect a single number as the answer. The queries are forbidden to provide the clues to the required domain-specific knowledge.
For the responses, the annotators are prompted to write the reasoning chains as detailed as possible, which help to explain and check the results. A single numeric answer is expected at the end for each query. For evaluation, we extract the numeric answer with regex and calculate the accuracy.

\paragraph{Quality Review}
To precisely assessing the calculation results of LLMs, we also carefully review the built dataset after the first round of annotation. We correct the improper selected documents and wrong calculations.

\paragraph{Extension by LLM}
Since the human labor is expensive, we adopt GPT-4 to extend the dataset scale given hand-written instances as examples for each case. We also apply manual quality review before adding the instances into the dataset.

\section{Method}

In this Section, we introduce the framework of KIPG, as illustrated in Figure \ref{fig:framework}.

\subsection{Knowledge-Intensive Programs}

Our pipeline works with a program generator LLM, denoted as $\theta_{G}$. $\theta_{G}$ writes knowledge-intensive programs for each article. To provide necessary information for variable extraction and knowledge understanding, we inject the description to the knowledge source, input/output parameters and program comment to the programs. Details about the above components are available in Appendix \ref{app:kip component}.

\begin{figure}
    \centering
    \includegraphics[width=\linewidth]{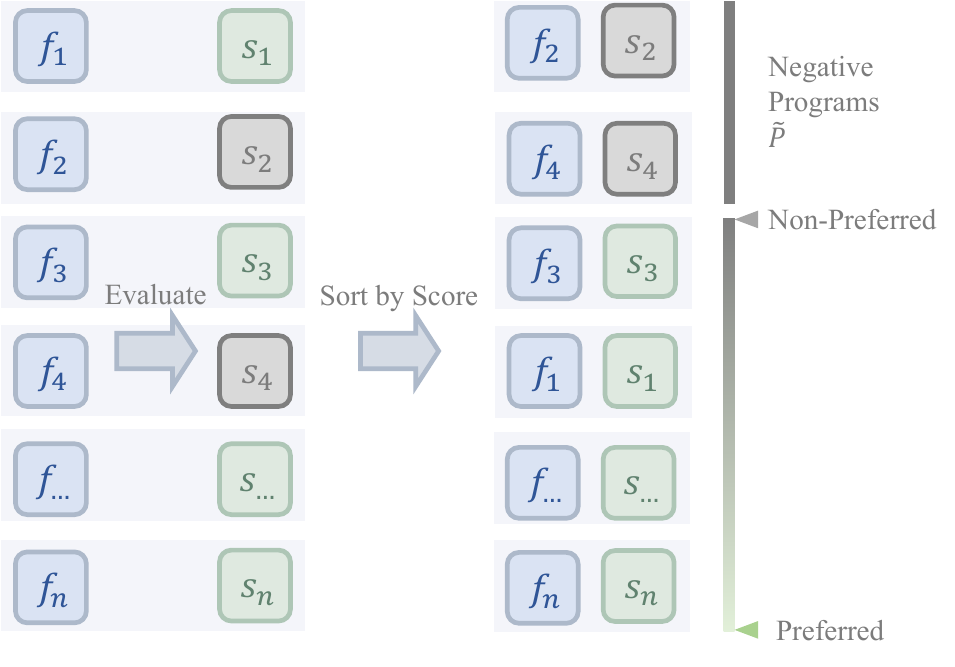}
    \caption{Ranking guideline of DPO data generation.}
    \label{fig:dpo data}
\end{figure}

\subsection{KIPG}

Given a query $Q$, the related domain-specific knowledge $D$ can be obtained by retrieval. Here, we focus on how to understand and strictly follow the instructions and rules with the documents as background knowledge. An extraction model, denoted as $\theta_{E}$, is adopted to extract the key variables from the query, according to the knowledge-intensive program $f$. The variables sever as input arguments to the program. Then, the program is executed by the system executor.
\begin{equation}
\begin{aligned}
    I &= \theta_{E}(Q, f)\\
    O &= f(I)
\end{aligned}
\end{equation}
The variables are packed together as $(I, O)$, which contain domain-specific information derived from the document and are appropriate for the case of $Q$. The response can be concluded by another model $\theta_{C}$ given the deduced information as
\begin{equation}
    Y = \theta_{C}(Q, D, (I, O))
\end{equation}
The original document $D$ helps $\theta_{C}$ to understand and review the variables, and the query $Q$ provides the calculation target of conclusion. Note that $(I, O)$  merely provide the answer derived from target document, not the direct answer to $Q$. For example, an article may stipulate the total amount to be reimbursed in the event of an accident, but the user inquires about the amount payable per month if the repayment is spread over 12 installments. In this way, it requires the $\theta_{C}$ to do post-calculation according to the query.

\begin{table*}[th]
    \centering
    \footnotesize
    \begin{tabular}{l|cccccccccc}
    \toprule
    \textbf{Type} & \textbf{Qwen-Max} & \textbf{Vanilla} & \textbf{EEDP} & \textbf{CoT} & \textbf{PoT} & \textbf{Algebraic} & \textbf{RaR} & \textbf{ICL} & \textbf{RCI} & \textbf{KIPG}\\
    \midrule
    \multicolumn{11}{c}{\textit{Qwen2-7B}}\\
    \midrule
    Compensation & 34.55 & 44.55 & 45.00 & 46.82 & 39.09 & 40.45 & 31.36 & 38.18 & 38.64 & \textbf{50.91}\\
    Tax & 73.15 & 60.65 & 62.04 & 64.35 & 30.56 & 59.72 & 40.28 & 57.87 & 60.19 & \textbf{73.15}\\
    Other Fees & 61.19 & 57.71 & 60.70 & 52.24 & 33.83 & 57.71 & 38.81 & 53.73 & 54.23 & \textbf{64.68}\\
    Penalties & 87.86 & 81.91 & 90.92 & 86.06 & 71.42 & 72.04 & 62.18 & 82.54 & 83.87 & \textbf{92.17}\\
    Traffic Violations & 62.50 & 48.53 & 62.50 & 55.88 & 45.59 & 55.15 & 38.97 & 50.00 & 54.41 & \textbf{68.38}\\
    AVG & 63.85 & 58.67 & 64.23 & 61.07 & 44.10 & 57.01 & 42.32 & 56.46 & 58.27 & \textbf{69.86}\\
    \midrule
    \multicolumn{11}{c}{\textit{Qwen2-72B}}\\
    \midrule
    Compensation & - & 49.55 & 52.73 & 50.00 & \textbf{57.27} & 52.27 & 28.18 & 51.36 & 51.36 & 56.82\\
    Tax & - & 68.98 & 75.00 & 76.12 & 75.46 & 61.57 & 39.35 & 70.00 & 73.61 & \textbf{77.78}\\
    Other Fees & - & 47.26 & 61.19 & 48.94 & 61.19 & 56.22 & 30.35 & 62.69 & 58.71 & \textbf{71.14}\\
    Penalties  & - & 77.76 & 93.81 & 83.16 & 90.37 & 83.32 & 54.78 & 90.27 & 86.03 & \textbf{95.69}\\
    Traffic Violations & - & 47.79 & 65.44 & 61.97 & 53.68 & 69.85 & 19.85 & 50.76 & 58.09 & \textbf{72.79}\\
    AVG  & - & 58.27 & 69.63 & 64.04 & 67.59 & 64.65 & 34.50 & 65.02 & 65.56 & \textbf{74.84}\\
    \bottomrule
    \end{tabular}
    \caption{Main results given oracle documents in legal domain. \textbf{The training dataset only includes the first 3 types, while the calculation problems of the ``Penalties'' and ``Traffic Violations'' are never seen during training.} We first conduct experiments on Qwen2-7B-Instruct, denoted as ``\textit{Qwen2-7B}'', which is additionally compared with ``Qwen-Max'' (200B parameters). Then we supplement the calculated variables from Qwen-7B directly to Qwen2-72B-Instruct, denoted as ``\textit{Qwen2-72B}''. ``Vanilla'' indicates directly using the document without any prompting or advanced skills.}
    \label{tab:oracle}
\end{table*}

\begin{table*}[th]
    \centering
    \footnotesize
    \begin{tabular}{l|ccccm{1.5cm}<{\centering}|c}
    \toprule
    \textbf{Method} & \textbf{Compensation} & \textbf{Tax} & \textbf{Other Fees} & \textbf{Penalties} & \textbf{Traffic Violations} & \textbf{AVG} \\
    \midrule
    \multicolumn{7}{c}{\textit{Zero/Few-Shot}}\\
    \midrule
        - & 18.64 & 33.33 & 30.85 & 31.95 & 24.26 & 27.81\\
        CoT & 19.55 & 27.78 & 29.35 & 33.36 &16.91 & 25.39\\
        ICL & 26.82 & 42.13 & 46.27 & 41.74 & 24.26 & 36.24\\
    \midrule
    \multicolumn{7}{c}{\textit{Supervised Fine-Tuning}}\\
    \midrule
        - & 26.36 & 37.50 & 26.87 & 36.81 & 18.38 & 29.18\\
        CoT & 25.45 & 40.74 & 28.36 & 36.41 & 16.91 & 29.57\\
        L1 & 28.18 & 31.94 & 32.33 & 27.88 & 19.85 & 28.44 \\
        L2 & 31.82 & 39.81 & 37.81 & 39.08 & 11.76 & 32.06 \\
        $\text{MixTraining}_{\text{DA}}$ & 36.82 & 47.69 & 51.74 & 45.42 & 21.32 & 40.60\\
        $\text{MixTraining}_{\text{GSM8K}}$ & 26.36 & 37.50 & 30.85 & 36.10 & 19.85 & 30.13 \\
        ICL & 27.27 & 38.89 & 39.30 & 44.24 & 17.65 & 33.47 \\
        SciAgent & 34.09 & 43.98 & 46.27 & 32.73 & 16.91 & 34.80 \\
        \midrule
        \multicolumn{7}{c}{\textit{Retrieve with LLM}}\\
        \midrule
        - & 31.36 & 40.28 & 40.80 & 48.71 & 16.91 & 35.61 \\
        Algebraic & 28.64 & 38.43 & 37.81 & 42.52 & 13.97 & 32.27 \\
        PoT & 30.45 & 25.93 & 22.89 & 44.01 & 14.71 & 27.60 \\
        CoT & 30.45 & 39.81 & 40.30 & 51.68 & 16.17 & 35.68 \\
        ICL & 28.64 & 43.52 & 39.80 & 48.08 & 17.65 & 35.54 \\
        RaR & 27.27 & 25.93 & 23.38 & 43.70 & 17.65 & 27.59 \\
        EEDP & 30.91 & 41.67 & 42.79 & 50.90 & 22.06 & 37.67 \\
        RCI & 32.73 & 37.50 & 30.35 & 46.99 & 22.06 & 33.93\\
        \midrule
        \multicolumn{7}{c}{\textit{Retrieve with SLM}}\\
        \midrule
        - & 41.82 & 58.33 & 56.22 & 66.09 & 21.32 & 48.76\\
        Algebraic & 37.27 & 56.94 & 56.22 & 61.55 & 24.26 & 47.25 \\
        PoT & 37.27 & 29.17 & 35.82 & 59.51 & 19.12 & 36.18\\
        CoT & 40.91 & 61.11 & 52.74 & 70.40 & 26.47 & 50.33\\
        ICL & 32.27 & 54.63 & 55.22 & 67.89 & 16.18 & 45.24\\
        RaR & 32.27 & 39.35 & 38.81 & 51.61 & 20.59 & 36.53\\
        EEDP & 46.82 & 59.26 & 62.19 & 75.10 & 23.53 & 53.38\\
        RCI & 40.00 & 56.48 & 52.74 & 68.36  & 26.47 & 48.81\\
        KIPG & \textbf{47.27} & \textbf{72.22} & \textbf{62.69} & \textbf{75.72} & \textbf{30.15} & \textbf{57.61}\\
        \bottomrule
    \end{tabular}
    \caption{Results without providing the label knowledge. We explore the potential internal domain-specific knowledge under zero/few-shot scenarios. We also try different training source and skills (``Supervised Fine-Tuning''). Additionally, we turn to RAG considering the LLM itself and language model with smaller scale as the retriever respectively. Detailed baselines are available in Appendix \ref{app:baseline}.}
    \label{tab:rag}
\end{table*}

\subsection{Code Generation}

To improve the logic consistency with the original documents, we propose to enhance $\theta_{G}$ with iterative Direct Preference Optimization (DPO) \cite{rafailov2024directpreferenceoptimizationlanguage}. Preference alignment is studied across all time, hindered by the fuzzy standard of ``helpful, useful, harmless'' responses. In our task, we are also facing the challenge of identifying the best programs, since even the programs generated by GPT-4 may also be problematic. Fortunately, we notice that evaluating the correctness of the generated programs is a rather simple task without demand of human labor, different from traditional preference alignment.

The ranking guideline of DPO data generation process is illustrated in Figure \ref{fig:dpo data}. For each article, we sample $n$ different programs with diverse beam search \cite{vijayakumar2018diversebeamsearchdecoding}, denoted as $P = [f_1, f_2, \dots, f_n]$. The programs are evaluated by KIPG respectively. We calculate the correctness when using a program $f_i$ to solve all queries requiring the corresponding document, and the score is denoted as $s_i$. There are also potential programs failed to be executed, whose scores are assigned as -1, and the program set is called negative programs, represented by $\tilde{P} \subseteq P$. We filter out $\tilde{P}$ by the following aspects: 1) If $f_i$ can not be executed and throw an runtime exception, then $f_i \in \tilde{P}$. 2) If the parameter definitions in the comment of $f_i$ are fuzzy, then $f_i \in \tilde{P}$. 3) If $f_i$ contains words like ``assuming'', it may generate hallucinations, fabricating content that does not exist, then $f_i \in \tilde{P}$.

All valid programs are sorted by the scores in ascending order. Then all later programs are more preferred than previous programs. Besides, executable programs are more preferred than negative programs. Gathering each pair of programs, we can build the dataset $D_G$, then train the code generator $\theta_G$ with DPO loss:
\begin{equation}
\begin{aligned}
    \mathcal{L}_{G} = &-\mathbb{E}_{(d, p_w, p_l) \sim D_G} \left[\log \sigma \left(\beta \log \frac{\theta_G\left(p_w|d\right)}{\theta_{\text{ref}}\left(p_w|d\right)} \right. \right. \\
    &\left. \left. - \beta \log \frac{\theta_G\left(p_l|d\right)}{\theta_{\text{ref}}\left(p_l|d\right)} \right) \right]
\end{aligned}
\end{equation}
where $\left \langle p_w, p_l \right \rangle$ represents a pair of positive program and negative program.

By iterative inference with new parameters, KIPG gets different groups of programs and scores, consequently constructs new training data. Thus the overall accuracy improves step by step.

\subsection{Initialization}

From experiments, we find that vanilla extraction LLM $\theta_E$ struggles to follow the extraction instruction. Mostly, it outputs the extraction in random formats, hindering the following usage of the variables. Hence, we initialize $\theta_E$ with supervised fine-tuning on extraction instances generated by GPT-4.

\section{Experiments}

We have conducted extensive experiments comparing KIPG with several competitive baselines, including vanilla LLMs \cite{yang2024qwen2technicalreport}, CoT \cite{wei2023chainofthoughtpromptingelicitsreasoning}, PoT \cite{chen2023programthoughtspromptingdisentangling}, Algebraic \cite{imani-etal-2023-mathprompter}, InContext-Learning \cite{dong2024surveyincontextlearning}, SciAgent \cite{ma2024sciagenttoolaugmentedlanguagemodels}, RaR \cite{deng2024rephraserespondletlarge}, RCI \cite{kim2023languagemodelssolvecomputer}, EEDP \cite{eedp}, and Qwen-Max\footnote{\url{https://huggingface.co/spaces/Qwen/Qwen-Max-0428}} with 200B parameters. We also try training techniques with different training datasets. Detailed baseline descriptions can be found in Appendix \ref{app:baseline}. We mainly report the accuracy in our experiments, calculated by regex from the sentences\footnote{\url{https://github.com/QwenLM/Qwen/blob/main/eval/evaluate\_chat\_gsm8k.py}}.

\subsection{Main Results}

\begin{table}[th]
    \centering
    \footnotesize
    \begin{tabular}{l|ccc|c}
    \toprule
    \textbf{Method} & \textbf{Indicator} & \textbf{Term} & \textbf{Medicine} & \textbf{AVG} \\
    \midrule
    Vanilla &44.20 & 65.36 & 51.08 & 53.55  \\
    CoT & 45.59 & 63.40 & 57.55 & 55.51 \\
    PoT & 36.76 & 52.29 & 48.20 & 45.75 \\
    Algebraic & 69.93 & 52.78 & 55.40 & 52.56 \\
    RaR & 7.35 & 26.80 & 23.74 & 19.30 \\
    RCI & 51.47 & 58.17 & 51.80 & 53.81 \\
    EEDP & 48.53 & 71.90 & 61.87 & 60.77 \\
    KIPG & \textbf{52.94} & \textbf{74.51} & \textbf{64.75} & \textbf{64.07}\\
    \bottomrule
    \end{tabular}
    \caption{Cross domain performance given the oracle context.}
    \label{tab:medical}
\end{table}

\begin{table}[t]
    \centering
    \footnotesize
    \begin{tabular}{l|m{0.6cm}<{\centering}m{0.6cm}<{\centering}m{0.8cm}<{\centering}m{0.6cm}<{\centering}m{0.9cm}<{\centering}}
    \toprule
        \textbf{Type} & \textbf{GPT-4} & \textbf{Code Qwen} & \textbf{Vanilla} & \textbf{SFT} & \textbf{Ours} \\
        \midrule
        \scriptsize{Compensation} & 49.09 & 44.09 & 43.18 & \textbf{52.27} & 50.91\\
        Tax & 72.69 & 52.78 & 0.00 & 71.30 & \textbf{73.15} \\
        Other Fees & 49.75 & 53.23 & 51.74 & 52.74 & \textbf{64.68}\\
        Penalties  & 87.94 &  90.37 & 91.31 & \textbf{93.42} & 92.17\\
        \scriptsize{Traffic Violations} & 55.88 & 61.76 & 1.47 & 64.71 & \textbf{68.38}\\
        \midrule
        AVG & 63.07 & 60.45 & 37.54 & 66.89 & \textbf{69.86}\\
        \bottomrule
    \end{tabular}
    \caption{Comparison between different code generators during calculation. CodeQwen is a specific LLM developed for code relative tasks. ``Vanilla'' indicates the general Qwen2-7B-Instruct model. ``SFT'' indicates the Qwen model initialized with the GPT-4 generated codes, followed by DPO training.}
    \label{tab:code-model}
\end{table}

\paragraph{Oracle Context}
We report the main results \textbf{given the gold document in legal domain as the context} in Table \ref{tab:oracle}, i.e., the context article contains the necessary domain-specific knowledge for solving the problem. $\theta_G$ and $\theta_E$ are always Qwen2-7B models, while using 7B and 72B models as $\theta_C$ (the conclusion model) respectively.
\textbf{1)} When we consider \textit{Qwen2-7B-Instruct} as $\theta_C$, the knowledge-intensive programs significantly improve the overall performance under all types of cases. Especially, KIPG on 7B model achieves equal or higher accuracy than Qwen-Max. The average accuracy is 69.86\%, which is 8.77\% relatively higher than the best baseline, EEDP. \textbf{2)} Then we prompt \textit{Qwen2-72B-Instruct} to perform as $\theta_C$. We omit the comparison with Qwen-Max under this setting. The accuracy of KIPG is mostly the highest over other methods. KIPG outperforms the baselines by a large margin. Additionally, by comparing the different model-scales, the 72B model is more skilled at leveraging the calculated domain-specific variables to generate the answers obviously. The average accuracy improves from 69.86\% to 74.84\%. It indicates the knowledge-intensive programs from smaller LLMs also benefit larger LLMs. \textbf{3)} Although KIPG is never trained on calculation cases of ``Penalties'' or ``Traffic Violations'', it also achieves the best results, indicating there are common features to exploit to solve the domain-specific calculation problems.

\paragraph{Without Oracle Context}
We conduct experiments without providing the golden knowledge under this setting. We prompt the models to either \textit{recall relative rules with inner knowledge} (``zero/few-shot'' and ``supervised fine-tuning'') or \textit{retrieve the most helpful document with RAG techniques} (``retriever with LLM'' and ``SLM'' such as BERT). The results are listed in Table \ref{tab:rag}. 
\textbf{1)} It exhibits better overall performance when retrieving the label documents using SLM than LLM. KIPG shows the highest accuracy for all types of cases. For ``Tax'', KIPG outperforms the second best by 11.11\%. KIPG also achieves the highest average accuracy of 57.61\%.
\textbf{2)} Besides, fine-tuning improves the accuracy of calculation generally. RAG further helps the models to perform better by providing relative documents. We find that there are hallucinations when considering LLM as retriever.

\subsection{Generalization across Domain}

To investigate the cross domain generalization, we test KIPG on constructed medical calculation dataset after training in legal domain. The results are presented in Table \ref{tab:medical}, where ``Term'' indicates the terminologies and concepts in legal domain, which is relatively simple. Correspondingly, it is observed that most methods have achieved better performance on this type of cases. KIPG reaches the accuracy of 74.51\% on ``Term'' and 64.07\% on average, surpassing the second best baseline EEDP.

\subsection{Iterations}

\begin{figure}[t]
    \centering
    \includegraphics[width=\linewidth]{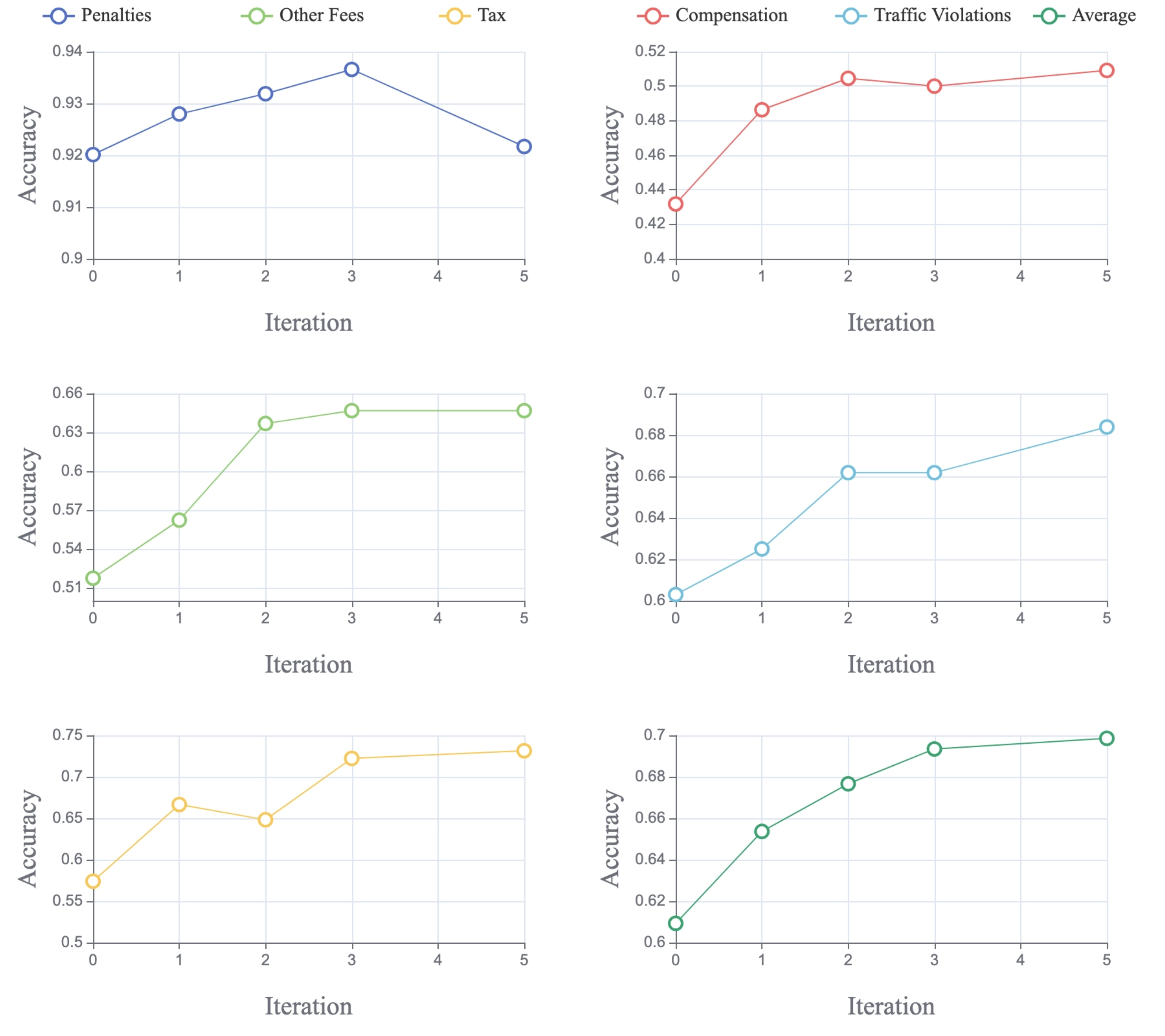}
    \caption{Accuracy over different training iterations under each type of cases.}
    \label{fig:over-iter}
\end{figure}

\begin{table}[t]
    \centering
    \footnotesize
    \begin{tabular}{l|ccc|c}
    \toprule
    \textbf{Method} & \textbf{Indicator} & \textbf{Term} & \textbf{Medicine} & \textbf{AVG} \\
    \midrule
    Vanilla &23.53 & 32.03 & 34.53 & 30.03  \\
    CoT & 26.47 & 24.84 & 38.13 & 29.81 \\
    PoT & 29.41 & 56.86 & 46.04 & 44.10 \\
    Algebraic & 26.47 & 30.72 & 43.88 & 33.69 \\
    RaR & 35.29 & 28.10 & 26.62 & 30.00 \\
    RCI & 32.35 & 45.10 & 48.92 & 42.12 \\
    EEDP & 32.35 & 51.63 & 36.69 & 40.22 \\
    KIPG & \textbf{39.71} & \textbf{65.36} & \textbf{51.80} & \textbf{52.29}\\
    \bottomrule
    \end{tabular}
    \caption{Cross domain performance in English conducted on Llama3 model with 8B parameters.}
    \label{tab:medical-en}
\end{table}

\begin{table*}[th]
    \centering
    \footnotesize
    \begin{tabular}{l|ccccc|c}
    \toprule
    \textbf{Method} & \textbf{Compensation} & \textbf{Tax} & \textbf{Other Fees} & \textbf{Penalties} & \textbf{Traffic Violations} & \textbf{AVG} \\
    \midrule
    KIPG & 50.45 & \textbf{64.81} & \textbf{63.68} & \textbf{93.19} & \textbf{66.18} & \textbf{67.66}\\
    \quad w/o init & 46.82{\tiny(-3.63)} & 63.43{\tiny(-1.38)} & 59.70{\tiny(-3.98)} & 92.95{\tiny(-0.24)} & 66.18{\tiny(0.00)} & 65.82{\tiny(-1.84)}\\
    \quad w/o syntax & \textbf{51.82{\tiny(+1.37)}} & 54.17{\tiny(-10.64)} & 54.73{\tiny(-8.95)} & 92.09{\tiny(-1.10)} & 63.97{\tiny(-2.21)} & 63.36{\tiny(-4.30)} \\
    \quad w/o correctness & 49.55{\tiny(-0.90)} & 55.56{\tiny(-9.25)} & 53.73{\tiny(-9.95)} & 92.48{\tiny(-0.71)} & 65.44{\tiny(-0.74)} & 63.35{\tiny(-4.31)} \\
    \quad w/o training & 43.18{\tiny(-7.27)} & 57.47{\tiny(-7.34)} & 51.74{\tiny(-11.94)} & 92.01{\tiny(-1.18)} & 60.29{\tiny(-5.89)} & 60.93{\tiny(-6.73)} \\
    \bottomrule
    \end{tabular}
    \caption{Ablation study at the second training iteration. ``w/o init'' indicates removing the initialization of extraction LLM. ``w/o syntax'' and ``w/o correctness'' indicate removing the syntax review and correctness review respectively during the DPO data ranking. ``w/o training'' indicates the setting without training $\theta_G$.}
    \label{tab:ablation}
\end{table*}

We illustrate the accuracy changes over different iterations in Figure \ref{fig:over-iter}. Given that the types of ``Other Fees'', ``Tax'' and ``Compensation'' appear in the training dataset, their accuracy improves rapidly to a high level. While for ``Traffic Violations'', there is a significant optimization at iteration 5. But the iteration also leads to reduction sometimes especially for ``Penalties'', which is unobserved during training. Generally, the accuracy presents an ascending trend with the iteration increasing.

\subsection{Code Models}

We also compare different code generators within KIPG inference framework in Table \ref{tab:code-model}. Although GPT-4 is the largest model, it is not good at generating the knowledge-intensive programs in legal domain. CodeQwen\footnote{\url{https://qwenlm.github.io/blog/codeqwen1.5}} is specifically developed for code relative tasks, while the average accuracy is only 60.43\%. From the above analysis, it can be seen that when it comes to domain knowledge, the generation mode of knowledge-intensive programs is different from that of ordinary codes. Our proposed training process achieves the best results on most types of cases, and it is better for some types when initializing $\theta_G$ by SFT.

\subsection{Different Language and Base LLM}

To verify the effectiveness of KIPG on different base LLMs and languages, we train Llama3-8B-Instruct \cite{llama3modelcard} on legal dataset in English and test it on translated English medical-domain dataset. The results are listed in Table \ref{tab:medical-en}. Surprisingly, PoT exhibits outstanding performance over other baseline methods under this setting, although it struggles from runtime errors in legal domain. KIPG has significant advantage among all approaches, with 8.19\% margin over PoT in average.

\subsection{Ablation Study}
The ablation study is shown in Table \ref{tab:ablation}. We list the results after \textbf{the second iteration}. KIPG exhibits the best overall accuracy, while removing the initialization of $\theta_E$ slightly reduces the performance from 67.66\% to 65.82\%. Then, it seems that code review on syntax and correctness have the same overall significance, while the accuracy of ``compensation'' is higher when ignoring syntax during DPO ranking. By removing training $\theta_G$, the accuracy is not satisfactory, with the dramatic reduction of 11.94\% on ``Other Fees'', and 6.73\% accuracy drop on average.

\subsection{Logical Complexity}

\begin{figure}[th]
    \centering
    \includegraphics[width=\linewidth]{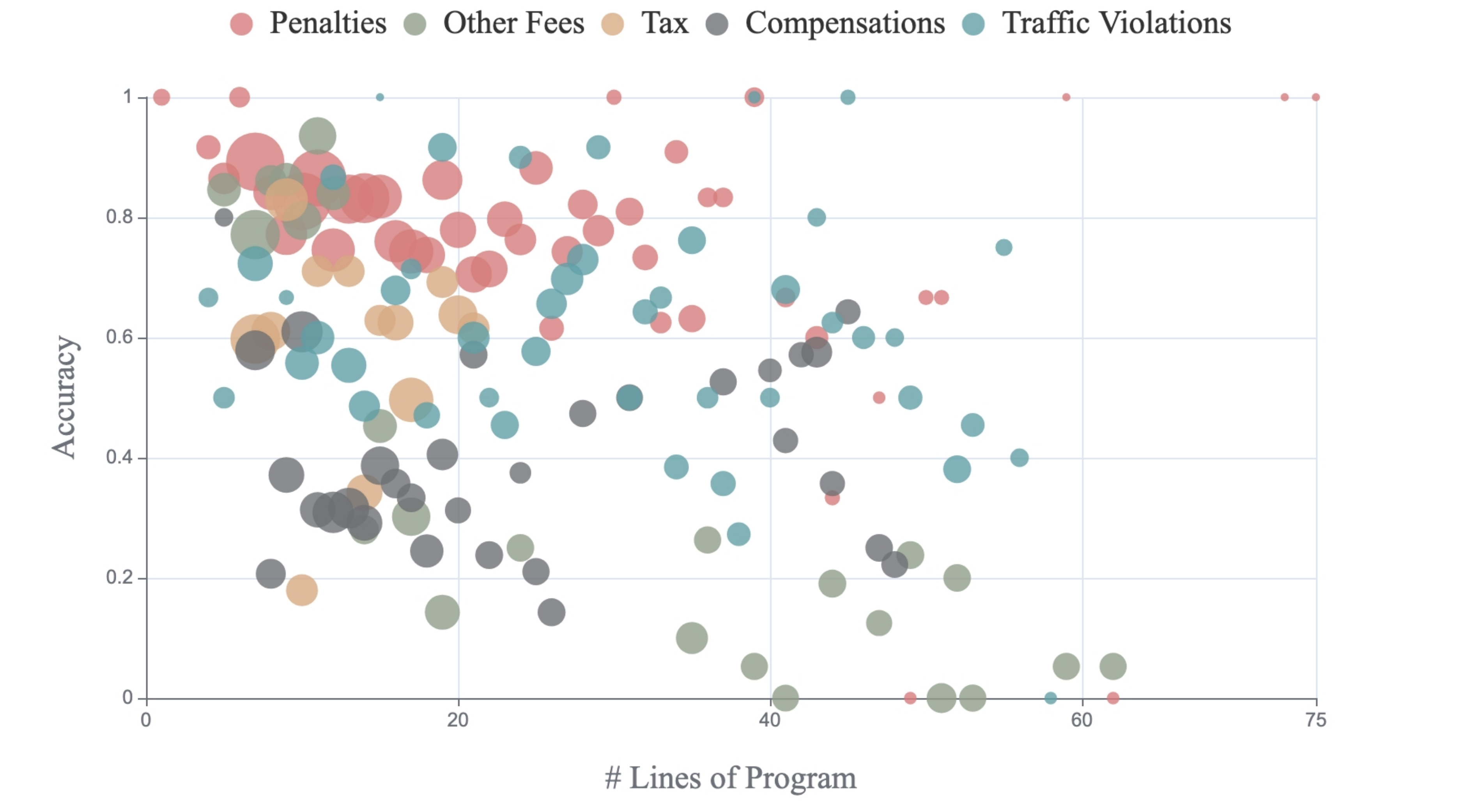}
    \caption{Number of lines in the programs versus the program accuracy under different types. The point size indicates the number of such programs.}
    \label{fig:distribution}
\end{figure}

We investigate the distribution of the accuracy versus the complexity of the documents and programs. Intuitively, a document describing complex rules and instructions corresponds to a program with more lines. Thus we utilize the number of program lines as a simplified indicator of the complexity of document. The distribution is illustrated in Figure \ref{fig:distribution}. In general, more complex the document is, there are more lines in the program, and the corresponding accuracy is lower. Thus the plot shows a descending trend as the number of lines increases. Additionally, it is reasonable that the type of ``Other Fees'' covers a wide range along the x-axis, since it contains instructions of various fees. The programs of ``Penalties'' mostly scattered on the left-top section, which explains why its overall accuracy is relatively high.

\section{Conclusion}

In this paper, we present the task of domain-specific calculation involving knowledge of complex rules and conditions, and the usage of knowledge-intensive programs. Additionally, we also introduce the framework named KIPG to solve the calculation task. We construct datasets in legal and medical domains with human labor and LLMs in a carefully designed pipeline. Empirical results illustrate the superiority of our method from multiple perspectives. KIPG outperforms other baselines with different base LLMs and different scales, in both Chinese and English.

\section*{Limitations}
In this paper, we propose the framework of KIPG to solve the domain-specific calculation problems involving documents describing complex instructions and conditions. Although KIPG presents the superiority given gold document, but our method have not directly optimized the retrieval. It can be seen that the average performance drop is more than 10\% without utilizing the label document. Obviously, all methods in our experiments dependent on RAG are all sensitive to the outcomes by retrieval. Optimizing the accuracy of retrieval is not our target in this paper, thus it may be remained for further studies.

\bibliography{acl_latex}

\begin{thebibliography}{28}
\providecommand{\natexlab}[1]{#1}

\bibitem[{AI@Meta(2024)}]{llama3modelcard}
AI@Meta. 2024.
\newblock \href {https://github.com/meta-llama/llama3/blob/main/MODEL_CARD.md} {Llama 3 model card}.

\bibitem[{Anil et~al.(2023)Anil, Dai, Firat, Johnson, Lepikhin, Passos, Shakeri, Taropa, Bailey, Chen, Chu, Clark, Shafey, Huang, Meier-Hellstern, Mishra, Moreira, Omernick, Robinson, Ruder, Tay, Xiao, Xu, Zhang, Abrego, Ahn, Austin, Barham, Botha, Bradbury, Brahma, Brooks, Catasta, Cheng, Cherry, Choquette-Choo, Chowdhery, Crepy, Dave, Dehghani, Dev, Devlin, Díaz, Du, Dyer, Feinberg, Feng, Fienber, Freitag, Garcia, Gehrmann, Gonzalez, Gur-Ari, Hand, Hashemi, Hou, Howland, Hu, Hui, Hurwitz, Isard, Ittycheriah, Jagielski, Jia, Kenealy, Krikun, Kudugunta, Lan, Lee, Lee, Li, Li, Li, Li, Li, Lim, Lin, Liu, Liu, Maggioni, Mahendru, Maynez, Misra, Moussalem, Nado, Nham, Ni, Nystrom, Parrish, Pellat, Polacek, Polozov, Pope, Qiao, Reif, Richter, Riley, Ros, Roy, Saeta, Samuel, Shelby, Slone, Smilkov, So, Sohn, Tokumine, Valter, Vasudevan, Vodrahalli, Wang, Wang, Wang, Wang, Wieting, Wu, Xu, Xu, Xue, Yin, Yu, Zhang, Zheng, Zheng, Zhou, Zhou, Petrov, and Wu}]{anil2023palm2technicalreport}
Rohan Anil, Andrew~M. Dai, Orhan Firat, Melvin Johnson, Dmitry Lepikhin, Alexandre Passos, Siamak Shakeri, Emanuel Taropa, Paige Bailey, Zhifeng Chen, Eric Chu, Jonathan~H. Clark, Laurent~El Shafey, Yanping Huang, Kathy Meier-Hellstern, Gaurav Mishra, Erica Moreira, Mark Omernick, Kevin Robinson, Sebastian Ruder, Yi~Tay, Kefan Xiao, Yuanzhong Xu, Yujing Zhang, Gustavo~Hernandez Abrego, Junwhan Ahn, Jacob Austin, Paul Barham, Jan Botha, James Bradbury, Siddhartha Brahma, Kevin Brooks, Michele Catasta, Yong Cheng, Colin Cherry, Christopher~A. Choquette-Choo, Aakanksha Chowdhery, Clément Crepy, Shachi Dave, Mostafa Dehghani, Sunipa Dev, Jacob Devlin, Mark Díaz, Nan Du, Ethan Dyer, Vlad Feinberg, Fangxiaoyu Feng, Vlad Fienber, Markus Freitag, Xavier Garcia, Sebastian Gehrmann, Lucas Gonzalez, Guy Gur-Ari, Steven Hand, Hadi Hashemi, Le~Hou, Joshua Howland, Andrea Hu, Jeffrey Hui, Jeremy Hurwitz, Michael Isard, Abe Ittycheriah, Matthew Jagielski, Wenhao Jia, Kathleen Kenealy, Maxim Krikun, Sneha Kudugunta, Chang
  Lan, Katherine Lee, Benjamin Lee, Eric Li, Music Li, Wei Li, YaGuang Li, Jian Li, Hyeontaek Lim, Hanzhao Lin, Zhongtao Liu, Frederick Liu, Marcello Maggioni, Aroma Mahendru, Joshua Maynez, Vedant Misra, Maysam Moussalem, Zachary Nado, John Nham, Eric Ni, Andrew Nystrom, Alicia Parrish, Marie Pellat, Martin Polacek, Alex Polozov, Reiner Pope, Siyuan Qiao, Emily Reif, Bryan Richter, Parker Riley, Alex~Castro Ros, Aurko Roy, Brennan Saeta, Rajkumar Samuel, Renee Shelby, Ambrose Slone, Daniel Smilkov, David~R. So, Daniel Sohn, Simon Tokumine, Dasha Valter, Vijay Vasudevan, Kiran Vodrahalli, Xuezhi Wang, Pidong Wang, Zirui Wang, Tao Wang, John Wieting, Yuhuai Wu, Kelvin Xu, Yunhan Xu, Linting Xue, Pengcheng Yin, Jiahui Yu, Qiao Zhang, Steven Zheng, Ce~Zheng, Weikang Zhou, Denny Zhou, Slav Petrov, and Yonghui Wu. 2023.
\newblock \href {https://arxiv.org/abs/2305.10403} {Palm 2 technical report}.
\newblock \emph{Preprint}, arXiv:2305.10403.

\bibitem[{Brown et~al.(2020)Brown, Mann, Ryder, Subbiah, Kaplan, Dhariwal, Neelakantan, Shyam, Sastry, Askell, Agarwal, Herbert-Voss, Krueger, Henighan, Child, Ramesh, Ziegler, Wu, Winter, Hesse, Chen, Sigler, Litwin, Gray, Chess, Clark, Berner, McCandlish, Radford, Sutskever, and Amodei}]{brown2020languagemodelsfewshotlearners}
Tom~B. Brown, Benjamin Mann, Nick Ryder, Melanie Subbiah, Jared Kaplan, Prafulla Dhariwal, Arvind Neelakantan, Pranav Shyam, Girish Sastry, Amanda Askell, Sandhini Agarwal, Ariel Herbert-Voss, Gretchen Krueger, Tom Henighan, Rewon Child, Aditya Ramesh, Daniel~M. Ziegler, Jeffrey Wu, Clemens Winter, Christopher Hesse, Mark Chen, Eric Sigler, Mateusz Litwin, Scott Gray, Benjamin Chess, Jack Clark, Christopher Berner, Sam McCandlish, Alec Radford, Ilya Sutskever, and Dario Amodei. 2020.
\newblock \href {https://arxiv.org/abs/2005.14165} {Language models are few-shot learners}.
\newblock \emph{Preprint}, arXiv:2005.14165.

\bibitem[{Chen et~al.(2023)Chen, Ma, Wang, and Cohen}]{chen2023programthoughtspromptingdisentangling}
Wenhu Chen, Xueguang Ma, Xinyi Wang, and William~W. Cohen. 2023.
\newblock \href {https://arxiv.org/abs/2211.12588} {Program of thoughts prompting: Disentangling computation from reasoning for numerical reasoning tasks}.
\newblock \emph{Preprint}, arXiv:2211.12588.

\bibitem[{Cobbe et~al.(2021)Cobbe, Kosaraju, Bavarian, Chen, Jun, Kaiser, Plappert, Tworek, Hilton, Nakano, Hesse, and Schulman}]{cobbe2021trainingverifierssolvemath}
Karl Cobbe, Vineet Kosaraju, Mohammad Bavarian, Mark Chen, Heewoo Jun, Lukasz Kaiser, Matthias Plappert, Jerry Tworek, Jacob Hilton, Reiichiro Nakano, Christopher Hesse, and John Schulman. 2021.
\newblock \href {https://arxiv.org/abs/2110.14168} {Training verifiers to solve math word problems}.
\newblock \emph{Preprint}, arXiv:2110.14168.

\bibitem[{Deng et~al.(2024)Deng, Zhang, Chen, and Gu}]{deng2024rephraserespondletlarge}
Yihe Deng, Weitong Zhang, Zixiang Chen, and Quanquan Gu. 2024.
\newblock \href {https://arxiv.org/abs/2311.04205} {Rephrase and respond: Let large language models ask better questions for themselves}.
\newblock \emph{Preprint}, arXiv:2311.04205.

\bibitem[{Dong et~al.(2024)Dong, Li, Dai, Zheng, Ma, Li, Xia, Xu, Wu, Chang, Sun, Li, and Sui}]{dong2024surveyincontextlearning}
Qingxiu Dong, Lei Li, Damai Dai, Ce~Zheng, Jingyuan Ma, Rui Li, Heming Xia, Jingjing Xu, Zhiyong Wu, Baobao Chang, Xu~Sun, Lei Li, and Zhifang Sui. 2024.
\newblock \href {https://arxiv.org/abs/2301.00234} {A survey on in-context learning}.
\newblock \emph{Preprint}, arXiv:2301.00234.

\bibitem[{Feng et~al.(2024)Feng, Qin, Huang, Zhang, and Lei}]{feng2024analyzingunderstandinglimitationsdpo}
Duanyu Feng, Bowen Qin, Chen Huang, Zheng Zhang, and Wenqiang Lei. 2024.
\newblock \href {https://arxiv.org/abs/2404.04626} {Towards analyzing and understanding the limitations of dpo: A theoretical perspective}.
\newblock \emph{Preprint}, arXiv:2404.04626.

\bibitem[{Hu et~al.(2021)Hu, Shen, Wallis, Allen-Zhu, Li, Wang, Wang, and Chen}]{hu2021loralowrankadaptationlarge}
Edward~J. Hu, Yelong Shen, Phillip Wallis, Zeyuan Allen-Zhu, Yuanzhi Li, Shean Wang, Lu~Wang, and Weizhu Chen. 2021.
\newblock \href {https://arxiv.org/abs/2106.09685} {Lora: Low-rank adaptation of large language models}.
\newblock \emph{Preprint}, arXiv:2106.09685.

\bibitem[{Imani et~al.(2023)Imani, Du, and Shrivastava}]{imani-etal-2023-mathprompter}
Shima Imani, Liang Du, and Harsh Shrivastava. 2023.
\newblock \href {https://doi.org/10.18653/v1/2023.acl-industry.4} {{M}ath{P}rompter: Mathematical reasoning using large language models}.
\newblock In \emph{Proceedings of the 61st Annual Meeting of the Association for Computational Linguistics (Volume 5: Industry Track)}, pages 37--42, Toronto, Canada. Association for Computational Linguistics.

\bibitem[{Ke et~al.(2023)Ke, Shao, Lin, Konishi, Kim, and Liu}]{ke2023continualpretraininglanguagemodels}
Zixuan Ke, Yijia Shao, Haowei Lin, Tatsuya Konishi, Gyuhak Kim, and Bing Liu. 2023.
\newblock \href {https://arxiv.org/abs/2302.03241} {Continual pre-training of language models}.
\newblock \emph{Preprint}, arXiv:2302.03241.

\bibitem[{Kim et~al.(2023)Kim, Baldi, and McAleer}]{kim2023languagemodelssolvecomputer}
Geunwoo Kim, Pierre Baldi, and Stephen McAleer. 2023.
\newblock \href {https://arxiv.org/abs/2303.17491} {Language models can solve computer tasks}.
\newblock \emph{Preprint}, arXiv:2303.17491.

\bibitem[{Ma et~al.(2024)Ma, Gou, Hao, Xu, Wang, Pan, Yang, Cao, Sun, Awadalla, and Chen}]{ma2024sciagenttoolaugmentedlanguagemodels}
Yubo Ma, Zhibin Gou, Junheng Hao, Ruochen Xu, Shuohang Wang, Liangming Pan, Yujiu Yang, Yixin Cao, Aixin Sun, Hany Awadalla, and Weizhu Chen. 2024.
\newblock \href {https://arxiv.org/abs/2402.11451} {Sciagent: Tool-augmented language models for scientific reasoning}.
\newblock \emph{Preprint}, arXiv:2402.11451.

\bibitem[{Nay et~al.(2023)Nay, Karamardian, Lawsky, Tao, Bhat, Jain, Lee, Choi, and Kasai}]{nay2023largelanguagemodelstax}
John~J. Nay, David Karamardian, Sarah~B. Lawsky, Wenting Tao, Meghana Bhat, Raghav Jain, Aaron~Travis Lee, Jonathan~H. Choi, and Jungo Kasai. 2023.
\newblock \href {https://arxiv.org/abs/2306.07075} {Large language models as tax attorneys: A case study in legal capabilities emergence}.
\newblock \emph{Preprint}, arXiv:2306.07075.

\bibitem[{Pal et~al.(2024)Pal, Karkhanis, Dooley, Roberts, Naidu, and White}]{pal2024smaugfixingfailuremodes}
Arka Pal, Deep Karkhanis, Samuel Dooley, Manley Roberts, Siddartha Naidu, and Colin White. 2024.
\newblock \href {https://arxiv.org/abs/2402.13228} {Smaug: Fixing failure modes of preference optimisation with dpo-positive}.
\newblock \emph{Preprint}, arXiv:2402.13228.

\bibitem[{Rafailov et~al.(2024)Rafailov, Sharma, Mitchell, Ermon, Manning, and Finn}]{rafailov2024directpreferenceoptimizationlanguage}
Rafael Rafailov, Archit Sharma, Eric Mitchell, Stefano Ermon, Christopher~D. Manning, and Chelsea Finn. 2024.
\newblock \href {https://arxiv.org/abs/2305.18290} {Direct preference optimization: Your language model is secretly a reward model}.
\newblock \emph{Preprint}, arXiv:2305.18290.

\bibitem[{Srivastava et~al.(2024)Srivastava, Malik, Gupta, Ganu, and Roth}]{eedp}
Pragya Srivastava, Manuj Malik, Vivek Gupta, Tanuja Ganu, and Dan Roth. 2024.
\newblock \href {https://arxiv.org/abs/2402.11194} {Evaluating llms' mathematical reasoning in financial document question answering}.
\newblock \emph{Preprint}, arXiv:2402.11194.

\bibitem[{Touvron et~al.(2023)Touvron, Martin, Stone, Albert, Almahairi, Babaei, Bashlykov, Batra, Bhargava, Bhosale, Bikel, Blecher, Ferrer, Chen, Cucurull, Esiobu, Fernandes, Fu, Fu, Fuller, Gao, Goswami, Goyal, Hartshorn, Hosseini, Hou, Inan, Kardas, Kerkez, Khabsa, Kloumann, Korenev, Koura, Lachaux, Lavril, Lee, Liskovich, Lu, Mao, Martinet, Mihaylov, Mishra, Molybog, Nie, Poulton, Reizenstein, Rungta, Saladi, Schelten, Silva, Smith, Subramanian, Tan, Tang, Taylor, Williams, Kuan, Xu, Yan, Zarov, Zhang, Fan, Kambadur, Narang, Rodriguez, Stojnic, Edunov, and Scialom}]{touvron2023llama2openfoundation}
Hugo Touvron, Louis Martin, Kevin Stone, Peter Albert, Amjad Almahairi, Yasmine Babaei, Nikolay Bashlykov, Soumya Batra, Prajjwal Bhargava, Shruti Bhosale, Dan Bikel, Lukas Blecher, Cristian~Canton Ferrer, Moya Chen, Guillem Cucurull, David Esiobu, Jude Fernandes, Jeremy Fu, Wenyin Fu, Brian Fuller, Cynthia Gao, Vedanuj Goswami, Naman Goyal, Anthony Hartshorn, Saghar Hosseini, Rui Hou, Hakan Inan, Marcin Kardas, Viktor Kerkez, Madian Khabsa, Isabel Kloumann, Artem Korenev, Punit~Singh Koura, Marie-Anne Lachaux, Thibaut Lavril, Jenya Lee, Diana Liskovich, Yinghai Lu, Yuning Mao, Xavier Martinet, Todor Mihaylov, Pushkar Mishra, Igor Molybog, Yixin Nie, Andrew Poulton, Jeremy Reizenstein, Rashi Rungta, Kalyan Saladi, Alan Schelten, Ruan Silva, Eric~Michael Smith, Ranjan Subramanian, Xiaoqing~Ellen Tan, Binh Tang, Ross Taylor, Adina Williams, Jian~Xiang Kuan, Puxin Xu, Zheng Yan, Iliyan Zarov, Yuchen Zhang, Angela Fan, Melanie Kambadur, Sharan Narang, Aurelien Rodriguez, Robert Stojnic, Sergey Edunov, and Thomas
  Scialom. 2023.
\newblock \href {https://arxiv.org/abs/2307.09288} {Llama 2: Open foundation and fine-tuned chat models}.
\newblock \emph{Preprint}, arXiv:2307.09288.

\bibitem[{Vijayakumar et~al.(2018)Vijayakumar, Cogswell, Selvaraju, Sun, Lee, Crandall, and Batra}]{vijayakumar2018diversebeamsearchdecoding}
Ashwin~K Vijayakumar, Michael Cogswell, Ramprasath~R. Selvaraju, Qing Sun, Stefan Lee, David Crandall, and Dhruv Batra. 2018.
\newblock \href {https://arxiv.org/abs/1610.02424} {Diverse beam search: Decoding diverse solutions from neural sequence models}.
\newblock \emph{Preprint}, arXiv:1610.02424.

\bibitem[{Wei et~al.(2023)Wei, Wang, Schuurmans, Bosma, Ichter, Xia, Chi, Le, and Zhou}]{wei2023chainofthoughtpromptingelicitsreasoning}
Jason Wei, Xuezhi Wang, Dale Schuurmans, Maarten Bosma, Brian Ichter, Fei Xia, Ed~Chi, Quoc Le, and Denny Zhou. 2023.
\newblock \href {https://arxiv.org/abs/2201.11903} {Chain-of-thought prompting elicits reasoning in large language models}.
\newblock \emph{Preprint}, arXiv:2201.11903.

\bibitem[{Yang et~al.(2023{\natexlab{a}})Yang, Xiao, Wang, Zhang, Bian, Yin, Lv, Pan, Wang, Yan, Yang, Deng, Wang, Liu, Ai, Dong, Zhao, Xu, Sun, Zhang, Liu, Ji, Xie, Dai, Fang, Su, Song, Liu, Ru, Ma, Wang, Liu, Lin, Nie, Guo, Sun, Zhang, Li, Li, Cheng, Chen, Zeng, Wang, Chen, Men, Yu, Pan, Shen, Wang, Li, Jiang, Gao, Zhang, Zhou, and Wu}]{yang2023baichuan2openlargescale}
Aiyuan Yang, Bin Xiao, Bingning Wang, Borong Zhang, Ce~Bian, Chao Yin, Chenxu Lv, Da~Pan, Dian Wang, Dong Yan, Fan Yang, Fei Deng, Feng Wang, Feng Liu, Guangwei Ai, Guosheng Dong, Haizhou Zhao, Hang Xu, Haoze Sun, Hongda Zhang, Hui Liu, Jiaming Ji, Jian Xie, JunTao Dai, Kun Fang, Lei Su, Liang Song, Lifeng Liu, Liyun Ru, Luyao Ma, Mang Wang, Mickel Liu, MingAn Lin, Nuolan Nie, Peidong Guo, Ruiyang Sun, Tao Zhang, Tianpeng Li, Tianyu Li, Wei Cheng, Weipeng Chen, Xiangrong Zeng, Xiaochuan Wang, Xiaoxi Chen, Xin Men, Xin Yu, Xuehai Pan, Yanjun Shen, Yiding Wang, Yiyu Li, Youxin Jiang, Yuchen Gao, Yupeng Zhang, Zenan Zhou, and Zhiying Wu. 2023{\natexlab{a}}.
\newblock \href {https://arxiv.org/abs/2309.10305} {Baichuan 2: Open large-scale language models}.
\newblock \emph{Preprint}, arXiv:2309.10305.

\bibitem[{Yang et~al.(2024)Yang, Yang, Hui, Zheng, Yu, Zhou, Li, Li, Liu, Huang, Dong, Wei, Lin, Tang, Wang, Yang, Tu, Zhang, Ma, Yang, Xu, Zhou, Bai, He, Lin, Dang, Lu, Chen, Yang, Li, Xue, Ni, Zhang, Wang, Peng, Men, Gao, Lin, Wang, Bai, Tan, Zhu, Li, Liu, Ge, Deng, Zhou, Ren, Zhang, Wei, Ren, Liu, Fan, Yao, Zhang, Wan, Chu, Liu, Cui, Zhang, Guo, and Fan}]{yang2024qwen2technicalreport}
An~Yang, Baosong Yang, Binyuan Hui, Bo~Zheng, Bowen Yu, Chang Zhou, Chengpeng Li, Chengyuan Li, Dayiheng Liu, Fei Huang, Guanting Dong, Haoran Wei, Huan Lin, Jialong Tang, Jialin Wang, Jian Yang, Jianhong Tu, Jianwei Zhang, Jianxin Ma, Jianxin Yang, Jin Xu, Jingren Zhou, Jinze Bai, Jinzheng He, Junyang Lin, Kai Dang, Keming Lu, Keqin Chen, Kexin Yang, Mei Li, Mingfeng Xue, Na~Ni, Pei Zhang, Peng Wang, Ru~Peng, Rui Men, Ruize Gao, Runji Lin, Shijie Wang, Shuai Bai, Sinan Tan, Tianhang Zhu, Tianhao Li, Tianyu Liu, Wenbin Ge, Xiaodong Deng, Xiaohuan Zhou, Xingzhang Ren, Xinyu Zhang, Xipin Wei, Xuancheng Ren, Xuejing Liu, Yang Fan, Yang Yao, Yichang Zhang, Yu~Wan, Yunfei Chu, Yuqiong Liu, Zeyu Cui, Zhenru Zhang, Zhifang Guo, and Zhihao Fan. 2024.
\newblock \href {https://arxiv.org/abs/2407.10671} {Qwen2 technical report}.
\newblock \emph{Preprint}, arXiv:2407.10671.

\bibitem[{Yang et~al.(2023{\natexlab{b}})Yang, Liu, and Wang}]{yang2023fingptopensourcefinanciallarge}
Hongyang Yang, Xiao-Yang Liu, and Christina~Dan Wang. 2023{\natexlab{b}}.
\newblock \href {https://arxiv.org/abs/2306.06031} {Fingpt: Open-source financial large language models}.
\newblock \emph{Preprint}, arXiv:2306.06031.

\bibitem[{Yiquan et~al.()Yiquan, Yuhang, Yifei, Ang, Siying, and Kun}]{wisdomInterrogatory}
Wu~Yiquan, Liu Yuhang, Liu Yifei, Li~Ang, Zhou Siying, and Kuang Kun.
\newblock \href {https://github.com/zhihaiLLM/wisdomInterrogatory} {wisdominterrogatory}.
\newblock Available at GitHub.

\bibitem[{Zhang et~al.(2023)Zhang, Chen, Jiang, Yu, Chen, Li, Chen, Wu, Zhang, Xiao, Wan, Wang, and Li}]{zhang2023huatuogpttaminglanguagemodel}
Hongbo Zhang, Junying Chen, Feng Jiang, Fei Yu, Zhihong Chen, Jianquan Li, Guiming Chen, Xiangbo Wu, Zhiyi Zhang, Qingying Xiao, Xiang Wan, Benyou Wang, and Haizhou Li. 2023.
\newblock \href {https://arxiv.org/abs/2305.15075} {Huatuogpt, towards taming language model to be a doctor}.
\newblock \emph{Preprint}, arXiv:2305.15075.

\bibitem[{Zhao et~al.(2024)Zhao, Liu, Long, Zhang, Zhao, and Cohan}]{zhao2024financemathknowledgeintensivemathreasoning}
Yilun Zhao, Hongjun Liu, Yitao Long, Rui Zhang, Chen Zhao, and Arman Cohan. 2024.
\newblock \href {https://arxiv.org/abs/2311.09797} {Financemath: Knowledge-intensive math reasoning in finance domains}.
\newblock \emph{Preprint}, arXiv:2311.09797.

\bibitem[{Zhou et~al.(2024)Zhou, Shi, Song, Yang, Jin, Guo, and Li}]{zhou2024lawgptchineselegalknowledgeenhanced}
Zhi Zhou, Jiang-Xin Shi, Peng-Xiao Song, Xiao-Wen Yang, Yi-Xuan Jin, Lan-Zhe Guo, and Yu-Feng Li. 2024.
\newblock \href {https://arxiv.org/abs/2406.04614} {Lawgpt: A chinese legal knowledge-enhanced large language model}.
\newblock \emph{Preprint}, arXiv:2406.04614.

\bibitem[{Çağatay Yıldız et~al.(2024)Çağatay Yıldız, Ravichandran, Punia, Bethge, and Ermis}]{yıldız2024investigatingcontinualpretraininglarge}
Çağatay Yıldız, Nishaanth~Kanna Ravichandran, Prishruit Punia, Matthias Bethge, and Beyza Ermis. 2024.
\newblock \href {https://arxiv.org/abs/2402.17400} {Investigating continual pretraining in large language models: Insights and implications}.
\newblock \emph{Preprint}, arXiv:2402.17400.

\end{thebibliography}

\appendix

\section{Baselines and Setting}\label{app:baseline}

\paragraph{CoT} \citet{wei2023chainofthoughtpromptingelicitsreasoning} proposed CoT, which starts from ``Let's think step by step''.

\paragraph{ICL} In-Context Learning \cite{dong2024surveyincontextlearning} provides several examples within the contexts. For simplification, in our implementations, we provide 2 specific example of solving the task, ignoring the types of cases.

\paragraph{L1 and L2} We attempt to utilize the normalization methods to maintain the general capabilities during training on domain-specific corpus.

\paragraph{MixTraining} Training on the mixture of different data source is proved to be beneficial for continual pre-training and SFT. We tried the combination of legal articles with GPT-4 extended calculation problems (denoted as ``MixTraining$_{\text{DA}}$'') and GSM8K \cite{cobbe2021trainingverifierssolvemath} (denoted as ``MixTraining$_{\text{GSM8K}}$'').

\paragraph{SciAgent} \citet{ma2024sciagenttoolaugmentedlanguagemodels} introduced SciAgent to retrieve several well-prepared functions (represented by python programs). In our experiment, there are no ready tools, thus we prepare the programs of numeric calculation instead (such as sum, subtraction, multiplication, division, maximum and minimum).

\paragraph{Algebraic} Inspired by \citet{imani-etal-2023-mathprompter}, we utilize the mentioned algebraic formulation to replace the calculation by LLM itself.

\paragraph{PoT} \citet{chen2023programthoughtspromptingdisentangling} proposed PoT, which writes programs directly to solve the queries.

\paragraph{RaR} \citet{deng2024rephraserespondletlarge} allows LLMs to rephrase and expand questions and provide responses in a single prompt, named RaR.

\paragraph{EEDP} \citet{eedp} proposed to prompt the LLM in the order of Elicit, Extract, Decompose, Predict, to solve the calculation task involving domain-specific knowledge, and the technique is called EEDP.

\paragraph{RCI} \citet{kim2023languagemodelssolvecomputer} utilized LLMs to execute computer tasks guided by natural language using a simple prompting scheme where the agent Recursively Criticizes and Improves its output (RCI).

\paragraph{Oracle Context}
Under this setting, we provide the label document as the context to the LLMs, thus they can directly generate the solution from the knowledge. For the Qwen2-72B model, we don't train a brand new code generator. Instead, the variables extracted by Qwen2-7B model are directly given, to investigate that whether the outcomes from smaller model could benefit the inference of larger LLM.

\paragraph{Without Oracle Context} Under this setting, we assume that the label document is unobserved, then the LLMs have to recall the related knowledge either by internal parameters or retrieval from external sources. For SFT, we prepare legal article QA as the basic domain-specific corpus, based on which, we attempt to mix GSM8K and calculation instances generated by GPT-4 respectively into the training dataset. For ``Retrieve with LLM'', we adopt a LLM fine-tuned on article QA as the retriever, prompting it to recall the related article content given queries. For ``Retrieve with SLM'', we adopt a BERT model as the retriever to rank the articles according to embedding distances.

\paragraph{Cross Type}
For all scenarios involving training in legal domain, we keep the types of ``Penalties'' and ``Traffic Violations'' as the test types, which indicates that their calculation instances are unobserved during training. In this way, we can investigate the cross-type performance of the methods.

\paragraph{Cross Domain}
The medical dataset is kept for cross domain experiments. Specifically, we train the code generator in legal domain, then prompting it to generate the corresponding programs given documents in medical domain.

\paragraph{Llama in English}
To verify KIPG in English, we translate the related documents and instances into English with GPT-4, and develop KIPG based on Llama3-8B-Instruct. We also present the results of cross-domain performance in medical domain, since llama has rather limited knowledge about Chinese legal articles.

\section{Implementation Details}\label{app:implementation}
We use LoRA \cite{hu2021loralowrankadaptationlarge} for all training settings with rank 8. The learning rate is set to $5\times 10^{-5}$, and training batch size as 16. For each iteration of DPO, we set the epochs to 3 and $\beta$ to 0.1. The diverse beam search samples 8 programs for each document. We adopt Qwen2-7B-Instruct as the base LLM under most of the settings and download the parameters from HuggingFace \footnote{\url{https://huggingface.co/Qwen/Qwen2-72B-Instruct}}. During training, the calculation instances only involve the types of ``Compensation'', ``Tax'' and ``Other Fees'', while the other two types ``Penalties'' and ``Traffic Violations'' solely presents their articles. We leave the medical domain dataset for cross-domain experiments. To verify the effectiveness in English, we also adopt Llama3-8B-Instruct\footnote{\url{https://huggingface.co/meta-llama/Meta-Llama-3-8B-Instruct}} for cross domain experiment. We adopt bf16 and train the models on 4 A100 80G GPUs.

\section{Efficiency Discussion}

\begin{figure}
    \centering
    \includegraphics[width=\linewidth]{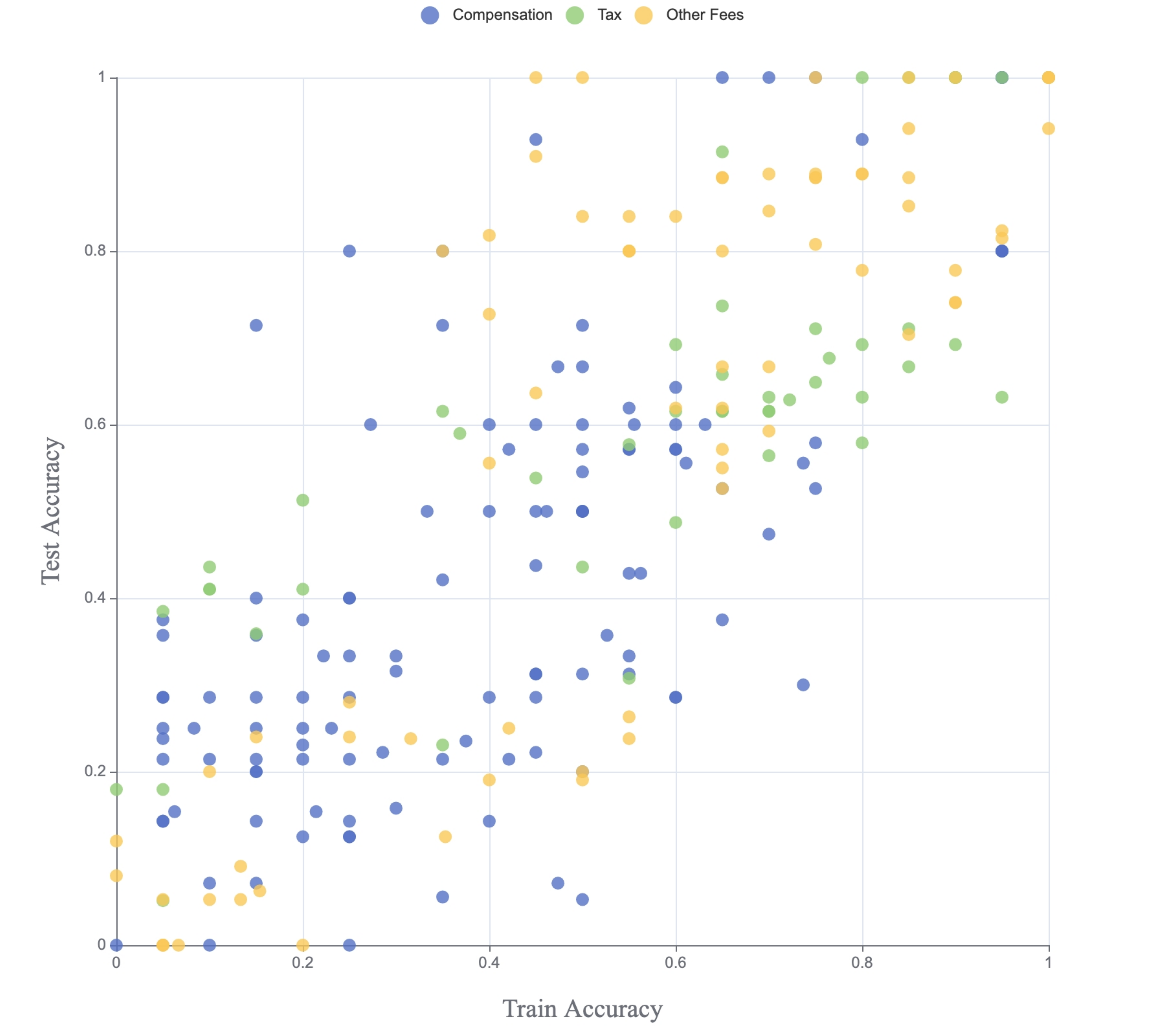}
    \caption{Accuracy of programs within the training set versus testing set.}
    \label{fig:efficiency}
\end{figure}

We discuss the efficiency of KIPG in this Section. First of all, we notice that the program generation is only applied in an offline manner. Thus the pre-generated programs can be directly called, given a query during inference. Additionally, we propose to filter the programs based on only a small set to reduce the calculation cost. Figure \ref{fig:efficiency} illustrates the distribution of programs accuracy in the large-scale testing set (2050 samples) and a small-scale training set (330 randomly selected samples) under different types of cases. It can be seen that there is a high consistency between the two accuracy. It indicates that the calculation cost can be greatly reduced by remaining only the programs that works well on the small-scale training set. Finally, KIPG doesn't change the behaviour of normal text generation, thus it is adaptable to the acceleration techniques, such as vLLM and Flash-Attention.

\section{Several Findings}

\paragraph{Larger beam size during training is important.}
We have tried different beam size during training, which means the scale of the exploration when generating knowledge-intensive programs. It is observed that larger beam size rapidly improves the conclusion performance with several iterations, and raises the upper limit after convergence. Larger exploration space directly increase the scale and diversity of the DPO training set, thus enhances the model performance.

\paragraph{Extraction LLM cannot be replaced by simple IE models.}
Extraction LLM $\theta_E$ extracts variables from the original query according to the comment of the programs. We attempt to replace LLM with a small language model for information extraction (IE), but it fails especially when the input variables requires simple general calculation and the text doesn't explicitly appear in the query. For example, the query presents the salary per year, while the program requires the salary per month. In this way, IE models cannot directly extract the correct value since the text span is not in the question.

\paragraph{Merging previous DPO instances does not help.}
We have also tried to merge the DPO data from previous iterations, to enlarge the scale of DPO training data and prevent over-fitting. However, the idea has only negative effects. We notice that the probabilities of both the chosen and rejected programs are descending, which may cause the model hard to generate the chosen one. \citet{feng2024analyzingunderstandinglimitationsdpo} and \citet{pal2024smaugfixingfailuremodes} also reported similar phenomenon. We assume the failure is caused by the nature of DPO.

\paragraph{Diverse documents help even without corresponding calculation examples.}
There are only 33 articles for our training types. We notice that by adding the articles of ``Compensation'' and ``Traffic Violations'' at the first iteration, the initial significantly improves. The direct consequence of introducing new articles is the larger DPO data about only program syntax, while it leads to the ultimate improvement in correctness surprisingly.

\section{Guidelines of Dataset Construction}
\label{app:guide dataset}

\paragraph{Knowledge Selection}
To improve the challenge of the task, we focus on practical domain knowledge, which mostly contains detailed instructions for different cases. 
For example, according to Article 13 of the Measures for the Payment of Litigation Costs, \textit{the part of property cases not exceeding 10,000 yuan shall be paid 50 yuan per case. The portion exceeding 10,000 yuan to 100,000 yuan shall be paid at 2.5\%. The portion exceeding 100,000 yuan to 150,000 yuan shall be paid at 2\%.} 
It is non-trivial for LLMs to strictly follow the complex knowledge document to perform calculation given queries, especially considering that it requires domain knowledge to identify the corresponding conditions.
For legal domain, we ask a team of lawyers with expertise in Chinese law to search satisfactory articles, given several examples. The articles cover a wide range of cases. For medical domain, we search for professional formulations and medication instructions on the Internet to ensure the correctness of the documents. Some statistics are provided in Table \ref{tab:data info}. 

\paragraph{Query and Responses}
In our dataset, queries are questions from users, and responses are the corresponding explanations including an answer from consultants. The human annotators are instructed to question in a manner of speaking as casual as possible. The queries should ask a specific question and expect a single number as the answer, without providing the clues to the required domain-specific knowledge.
For the responses, the annotators are prompted to write the reasoning as detailed as possible, which help to understand and check the results. A single numeric answer is expected at the end for each query. 
For the domain-specific documents describing a range of possible results, which is not rare, we turn to asking the maximum or minimum value in a general way. The results are format with 4 decimal places if they cannot be represented as integers. Additionally, we ask annotators to specify the unit of the answer, which is essential for evaluation.

\paragraph{Quality Review}
To precisely assessing the calculation results of LLMs, we also review the built dataset after the first round of annotation. Our review corrects the flaw including: 1) Incorrect question target, including asking multiple targets or requiring unrelated knowledge to the documents. 2) Wrong answers caused by incorrect calculation or low precision in intermediate variables. 3) Unclear explanations and reasoning. 4) Wrong citation to related articles.

\paragraph{Extension by LLM}
Since the human labor is expensive, we adopt GPT-4 to extend the dataset scale given hand-written instances as examples for each case. We provide detailed instructions and examples to GPT-4. We also specify the response with a explicit template ``\textit{According to \{knowledge\}, \{reasoning\}. \{Analyse the case.\} So the answer is \{answer\}.}'' We also apply data review before adding the instances into the dataset.

\section{Detailed Legal Documents}

\begin{table*}[th]
    \centering
    \footnotesize
    \begin{tabular}{lp{7cm}}
    \toprule
    \textbf{Type} & \textbf{SubType} \\
    \midrule
    \multirow{12}{*}{Compensation} & Funeral Allowance \\
    & Burial Expenses\\

& Medical Malpractice Compensation\\

& Work Injury Benefits\\

& Work-related Death Funeral Allowance\\
&Death Compensation\\

& Disability Compensation\\

& Economic Compensation\\

& Lost Wages\\

& Compensation\\

& Compensation Payment\\

& Non-work-related Death Funeral Allowance and Consolation Payments\\
    \midrule
    \multirow{11}{*}{Tax} & Personal Income Tax\\

&Taxable Income for Personal Income Tax\\

&Urban Maintenance and Construction Tax\\

&Stamp Tax\\

&Assessed Taxable Price of Taxable Vehicles\\

&Tobacco Leaf Tax\\

&Environmental Protection Tax\\

&Tax Arrears Penalty\\

&Cultivated Land Occupation Tax\\

&Vehicle Purchase Tax\\

&Interest on Debt During the Period of Delayed Performance\\
\midrule
\multirow{8}{*}{Other Fees} & Unemployment Insurance Premium \\

&Deposit\\

&Trade Union Funds\\

&Nursing Expenses\\

&Application Fee\\

&Dependent's Living Expenses\\

&Litigation Costs\\

&Preparation Fees\\
\midrule
    Penalties & Penalties\\
\midrule
    Traffic Violations & Traffic Violations\\
    \bottomrule
    \end{tabular}
    \caption{Types and sub-types in legal cases.}
    \label{tab:type list}
\end{table*}

Detailed types of legal cases are listed in Table \ref{tab:type list}.

\section{Components in Knowledge-Intensive Programs}\label{app:kip component}

\paragraph{Knowledge Source} We add the knowledge source to the start of the top comment, such as ``\textit{Calculate the compensation fees for personal injury in accordance with Article 7 of the Interpretation of the Supreme People's Court on Several Issues Concerning the Application of Law in the Trial of Cases Involving Compensation for Personal Injury.}''
\paragraph{Input and Output Arguments} In the top comment of the programs, we prompt the generator to supplement with detailed descriptions to the input and output arguments, including the definitions, units and data-types in Python language. Different from query/terminology-oriented programs, our programs return all potential outcomes that can be calculated from the document in a dictionary, instead of a scalar representing a single concept or the direct answer.
\paragraph{Comment Augmentation} To improve the logic consistency with the original document, we cite the sentences from the document before important calculation and ``if'' clauses. For example, the sentences start from ``\textit{The law states}'' or ``\textit{According to the law}'' for legal domain.


\section{Dataset Examples}

\begin{table*}[]
    \centering
    \footnotesize
    \begin{tabular}{c|p{5cm}|p{5cm}}
    \toprule
        Type & Query & Response \\
    \midrule
        Compensation & Zhang, a migrant worker, was hospitalized for 5 days due to his infringement, but he could not provide proof of fixed income or the average income of the last three years. It is known that the average salary of employees in the same or similar industries where Zhang lived in the previous year was 80,000 yuan/year. How much should Zhang get for lost work? & Zhang has no fixed income and cannot provide proof of the average income of the last three years, which is calculated according to the average salary of employees in the same or similar industries in the previous year. Therefore, Zhang deserves 80,000 yuan /365 days x 5 days =1,095.89 yuan.\\
        \midrule
        Penalties & A construction unit illegally built a small hydropower project on the Qinghai-Tibet Plateau, with a total investment of 5 million yuan. If the local people's government at or above the county level orders the construction to stop and restore to the original state, what is the maximum fine that the construction unit may face according to the law? & According to the fine range stipulated in Article 57 of the Qinghai-Tibet Plateau Ecological Protection Law of the People's Republic of China, the total investment amount of the construction project is determined to be 5 million yuan. According to laws and regulations, new small hydropower projects in violation of the provisions of this Law will be imposed a fine of not less than 1 percent but not more than 5 percent of the total investment of the construction project. Calculate the maximum fine: 5 million yuan ×5\% = 250,000 yuan.\\
        \midrule
        Other Fees & My wife and I are in the midst of divorce proceedings involving division of property. The total amount of our joint property is 3 million yuan. How much litigation fees should we pay in this case? & According to Article 13 (2) of the "Measures for the Payment of Litigation Costs", each divorce case pays 50 yuan to 300 yuan, if it involves the division of property and the total amount of property exceeds 200,000 yuan, it is paid in accordance with 0.5\%. In this case, the total amount of property is 3 million yuan, and the part exceeding 200,000 yuan is 2.8 million yuan. Therefore, it is necessary to pay 2.8 million yuan x 0.5\% = 1,400 yuan. With the minimum divorce case fee of 50 yuan, the total payment fee is 1400 yuan + 50 yuan = 1450 yuan.\\
    \bottomrule
    \end{tabular}
    \caption{Dataset Examples.}
    \label{tab:dataset example}
\end{table*}

We provide some examples of our constructed dataset in Table \ref{tab:dataset example}.

\section{Notations}

\begin{table}[th]
    \scriptsize
    \centering
    \begin{tabular}{l|p{5cm}}
    \toprule
    Notation & Description \\
    \midrule
        $\theta_G$ & The program generator. \\
        $\theta_E$ & The extraction model.\\
        $\theta_C$ & The conclusion model.\\
        $P = [f_1, f_2, \cdots, f_n]$ & List of programs sampled from $\theta_G$.\\
        $Q$ & The user query.\\
        $I, O$ & The input and output variables of a program.\\
        $\tilde{P} \subseteq P$ & The negative subset of $P$. Its programs are not executable because of errors.\\
        \bottomrule
    \end{tabular}
    \caption{Descriptions of the notations in this paper.}
    \label{tab:notations}
\end{table}

We clarify our notations in Table \ref{tab:notations}.



\end{document}